# Bayesian Functional Connectivity and Graph Convolutional Network for Working Memory Load Classification

**Harshini Gangapuram [1], Vidya Manian[1]**
[1]Department of Bioengineering, University of Puerto Rico, Mayaguez, PR 00682 USA
Corresponding author: Harshini Gangapuram (e-mail: harshini.gangapuram@upr.edu)

**ABSTRACT** Brain responses related to working memory originate from distinct brain areas and oscillate at different frequencies. EEG signals with high temporal correlation can effectively capture these responses. Therefore, estimating the functional connectivity of EEG for working memory protocols in different frequency bands plays a significant role in analyzing the brain dynamics with increasing memory and cognitive loads, which remains largely unexplored. The present study introduces a Bayesian structure learning algorithm to learn the functional connectivity of EEG in sensor space. Next, the functional connectivity graphs are taken as input to the graph convolutional network to classify the working memory loads. The intrasubject (subject-specific) classification performed on 154 subjects for six different verbal working memory loads produced the highest classification accuracy of 96% and average classification accuracy of 89%, outperforming state-of-the-art classification models proposed in the literature. Furthermore, the proposed Bayesian structure learning algorithm is compared with state-of-the-art functional connectivity estimation methods through intersubject and intrasubject statistical analysis of variance. The results also show that the alpha and theta bands have better classification accuracy than the beta band.

## I. INTRODUCTION

Working memory (WM) temporarily stores information while manipulating a given cognitive task [1]. The memory load imposed on the WM while performing a given cognitive task is called cognitive load. The brain has minimal capacity to store (hold) information while performing a cognitive task. Hence, WM declines with increasing cognitive load and results in performance breakdown, such as air traffic control, and low WM load leads to disengagement from the task [2], [3], [4]. Therefore, determining the ideal workload for an individual is essential in situations where individuals have to make decisions in real-time.

Classification of WM load is necessary to maintain work performance efficiency and diagnose disorders such as attention deficit hyperactivity disorder (ADHD) [5], bipolar disorders [6], schizophrenia[7], depression [8], mild cognitive impairment (MCI) [9], dementia, and Alzheimer's disease (AD)[10]. On the other hand, researchers demonstrated that emotional intelligence plays an essential role in performing WM tasks and classification of WM load can indicate stability and emotional intelligence [11].

There are many approaches to classifying WM load, such as subjective ratings, performance-based, behavioral, and physiological methods.

Physiological measurements such as heart rate variability (HRV) [12], hormone levels [13], galvanic skin response (GSR) [14], eye tracking systems [15], and measuring brain activity [16] provide direct response of the body for a given task. However, monitoring brain activity is considered one of the consistent physiological methods to assess WM load as it enables us to interface directly with the brain [17].

State-of-the-art methods extract different features, including frequency, time, spatial, and time-frequency domain features, from raw EEG signals and use them to classify the WM load. These methods, while advanced, often emphasize the local or regional properties (regions of interest) of the brain activity, potentially ignoring the global characteristics. This focus on specific areas may lead to a limited understanding of the complex and dynamic interactions that occur across all brain regions while performing a given cognitive task [18], [19]. Assessing such interactions provides relevant information about brain functioning between disease and control groups. "Brain connectivity" refers to the statistical inference of such interactions using neuroimaging modalities. The statistical inference depends on the brain sensing modality and the characteristics of the signal. Many statistical methods (connectivity metrics) have been proposed in the literature [19], [20], [21], [22], [23], [24], [25] and can be broadly classified into anatomical, functional, and effective connectivity. The anatomical connectivity corresponds to the white matter zones which represent the fiber tracts connecting the multiple brain regions [20]. Functional connectivity [21] infers the statistical dependencies among the time-series signals (temporal correlations) of brain activity. Finally, effective connectivity refers to the indirect or direct causal inference of one brain region on the other [22].

Brain connectivity studies aim to identify the correlated patterns of behavior and pathological phenotypes through statistical inferences. However, the existing methods focus on pairwise correlations such as in [26], [27], and with limited research in multivariate correlating edges [28], [29], [30], leading to biased connectivity patterns such as Pearson correlation. For example, if an edge (connection) exists between A and B brain regions, and an edge exists between B and C brain regions, imposing an edge between A and C is imposed, resulting in *type 1* errors (false positive) [31], [32]. In this case, an edge is perceived between two brain regions when there might not be any direct correlation. Partial correlation methods tried to reduce these errors, but biased connections still exist in the network structure [33]. Methods like graphical least absolute shrinkage and selection operator (GLASSO) regularize the elements in the connectivity (precision) matrix achieved through partial correlation through $l_1$-norm. This method provides sparse solutions by shrinking partial correlations to zero, introducing bias in the functional connectivity, and hence inducing *type 2* errors (false negative), where an actual connection exists between two brain regions [32], [34], [35]. Furthermore, these methods are limited to static functional connectivity ignoring the dynamic (temporal) connectivity of the brain while performing a given task.

While a few studies have achieved Bayesian functional connectivity using probabilistic graphical models, they are limited to a few studies estimating the functional connectivity in fMRI [36], [37], [38], [39] while the subjects are performing a cognitive task; therefore, suggesting more studies on brain functional connectivity need to be explored [40]. Estimating a Bayesian network has two phases: (i) determine graphical structure; and (ii) determine the parameters. Bayesian statistics are used significantly when addressing complex and uncertain data that involves expert and prior knowledge. Researchers determine that learning a network structure becomes NP-hard with increasing number of variables in the data [41].

Different Bayesian Structure learning (BSL) algorithms are proposed in the literature, which can be broadly classified into constraint-based, score-based, and hybrid. The score-based algorithms use a goodness-of-fit score that learns the data structure by iteratively maximizing the score, whereas constraint-based algorithms learn the dependence structure of data using conditional independence tests. The hybrid algorithms combine both methods. Of these algorithms, score-based methods are fast and produce efficient results. These algorithms include local search, genetic algorithms, simulated annealing, and greedy search; a thorough review is provided in [41], [42]. The score-based structure learning consists of two elements: (i) the search strategy to determine the path of adding/removing edges between the nodes of a graph; and (ii) the objective function (scoring method) to score the graph structures to identify the best graph that fits the data. Score-based structure learning aims to find the highest scoring graph among all the probabilities (all possible graphs) [41]. This paper proposes a score-based BSL to estimate the dynamic functional connectivity of the brain in sensor space. The Bayesian score has the following advantages.

- It considers the likelihood of the observed data given the network structure, which reflects how well the structure explains the data.
- It incorporates prior knowledge or beliefs about the conditional probability distributions of the variables, which can help to avoid overfitting and improve generalization.
- It penalizes complex network structures, such as those with many edges or cycles, by assigning lower scores to them. This helps to prevent overfitting and encourages simpler, more interpretable models.
- It can be computed efficiently using the Dirichlet and gamma functions, which are computationally tractable.

Therefore, it is a flexible method to learn the structure of Bayesian networks from data, considering both the likelihood of the data and prior knowledge about the network topology. Furthermore, we estimate the dynamic functional connectivity by performing BSL for each sliding window of raw EEG data corresponding to each trial. Therefore, it enables us to capture the subtle functional connectivity changes associated with WM loads. The algorithm is flexible in choosing different window sizes, allowing for customization based on the specifics of the EEG data.

Machine learning algorithms proposed in the literature, such as support vector machines (SVM) [43], [44], linear discriminant analysis (LDA) [45], k-nearest neighbors, decision trees and deep learning algorithms such as convolutional neural networks (CNNs) and LSTM [46] use EEG features to classify different cognitive tasks. These classification models perform better in image classification because of the regular structure of images. However, these methods cannot produce accurate results because of the graphical structure of the functional connectivity metrics, as graphs have distinctive non-Euclidean properties [47]. To completely utilize the information of functional connectivity networks, a graph convolutional neural network (GCN) has been used as an alternative to CNN. GCN can efficiently leverage the network structure and aggregate the nodes' information through convolutions of the neighborhoods. Therefore, GCNs have a superior performance in learning the structure of networks and can be used in a broader range of EEG classifications such as emotion recognition [48], driver state monitoring, AD, sleep stages, and seizure detection [49].

This paper aims to investigate the performance of the score-based BSL algorithm compared to different connectivity metrics proposed in the literature for WM functional connectivity estimation. Furthermore, a novel method is proposed to classify WM load using BSL functional connectivity features and the graph convolutional network (GCN) model. Low, medium, and high WM classes for two different cognitive loads (a total of six classes) are classified in this paper. The proposed method outperforms other classifiers and other combinations of functional connectivity metrics and classifiers. Also, it outperforms state-of-the-art WM load classifiers using functional connectivity metrics. Finally, intrasubject statistical analysis of BSL connectivity metrics is performed in alpha, beta, and theta frequency bands. The paper has the following contributions:

- The current study proposes a novel method to classify intrasubject (subject-specific) EEG dynamic functional connectivity features in sensor space based on the BSL algorithm and GCN model.
- The current study successfully outperforms the functional connectivity-based classification methods proposed in the literature with 96% accuracy.
- The current study successfully classifies the WM load corresponding to two cognitive tasks: manipulation and retention of letters (description of the dataset under section 2).
- The intrasubject and intersubject statistical analysis show that BSL produces consistent results in alpha, beta, and theta bands compared to state-of-the-art connectivity methods.

The rest of the paper is organized as follows: Section II presents the verbal working memory EEG dataset for WM loads. This section also presents the BSL methodology to estimate functional connectivity and the GCN model to classify the functional connectivity metrics. Section III presents the classification results and discussion of the proposed method in comparison with state-of-the-art methods. Section IV presents the conclusion.

## II. MATERIALS AND METHODS

The current study aims to propose Bayesian functional connectivity estimation and GCN classifier to perform WM load classification. The verbal WM dataset available in the OpenNeuro database is used in this study. The subjects performed verbal WM tasks for different memory and cognitive loads. A total of six loads are analyzed in this study. The artifacts of the dataset were removed by the authors while publishing the dataset through independent component analysis (ICA) [50]. First, the EEG dataset is bandpass filtered into alpha, beta, and theta bands. Next, EEG functional connectivity features are extracted using the BSL algorithm. Then, the functional connectivity features are classified using GCN. Finally, the results of the proposed method are compared with state-of-the-art methods. The methodology overview is presented in Figure 1.

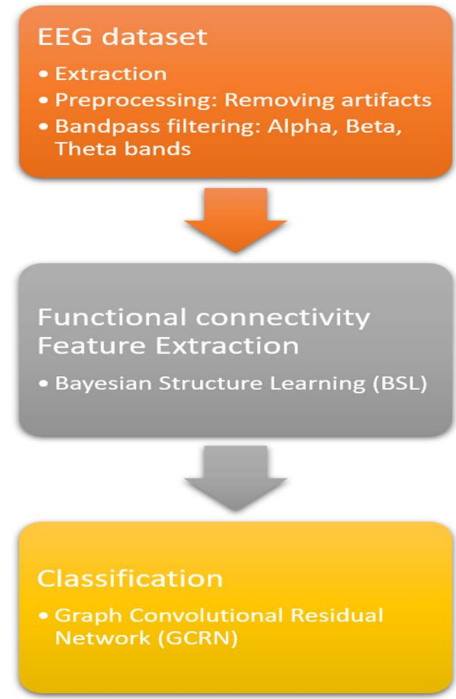

*Figure 1. Methodology overview*

### A. EEG data and preprocessing

EEF dataset of 156 subjects (mean age is 21) who participated in the WM task of retention and manipulation of the given alphabets is available in the OpenNeuro database [50], [51]. All the subjects were healthy with no history of mental disorders.

The alphabet is comprised of 1200 sets of letters that form no meaningful words and are presented in three loads: 5, 6, and 7 (400 sets for each load level). The trial begins with an exclamation mark for 200ms and a fixation cross for 3000ms. Then either "forward" or "alphabetical" words are displayed on the screen for 600ms to instruct the participants. The participants must retain letters for the "forward" task and manipulate letters for the "alphabetical" task. Next, the subjects must remember 5, 6, or 7 sets of letters displayed on the screen for 3000ms, and a delay period with a fixation cross is displayed for 6700ms. Then, a screen displays a letter randomly chosen from the previously displayed set and is presented on the screen with the sequential number of this letter with a letter-digit probe for 1000ms. The subject has to respond if the serial number of the letter is correct according to the retention or manipulation task. The subsequent trial started after 5000-5500ms.

Each task (a total of 6 tasks) had 20 consecutive trials. EEG data is collected from 19 EEG electrodes (channels) using Mitsar-EEG-202 amplifier that is a 10-20 system. These channels include Fp1, Fp2, F7, F3, Fz, F4, F8, T3, C3, Cz, C4, T4, T5, P3, Pz, P4, T6, O1, and O2. The sampling frequency is 500 Hz. The EEG artifacts are removed in two steps. First, to suppress the ocular activity artifacts, independent component analysis (AMICA algorithm) [52] was performed using the EEGLAB, a toolbox in MATLAB. Similarly, blinks and eye movements associated components were removed after visually exploring the data. The EEG epochs that contain noise and artifacts are recognized visually and discarded. The experimental paradigm and detailed description of the dataset is available in [50].

An average reference of all EEG channels is set so that the brain-specific fluctuations in electric potential are averaged to all EEG channels. Next, the Butterworth bandpass filter is used to extract three EEG frequency bands: theta (4-8 Hz), alpha (8-13 Hz), beta (15-20 Hz), and the connectivity metrics are computed in each frequency band. Finally, the data of two subjects (87 and 98) is eliminated because of discrepancies.

### B. Feature Extraction

#### I. BAYESIAN STRUCTURE LEARNING ALGORITHM

The steps for the BSL are as follows:
- Initialize a random network structure G.
- If the graph is not weakly connected, repeat the following steps for the specified number of iterations:
  - Select a pair of nodes (EEG channels) $(i, j)$ to modify such that $i \neq j$ and there is an edge between the nodes (EEG channels) $i$ and $j$ in the current network structure.
  - Evaluate the change in BIC that results from the modification:
    - Calculate the Bayesian score for the current network structure.
    - Modify the network structure by adding/removing the edge between the nodes $i$ and $j$.
  - Calculate the BIC for the modified network structure and estimate the change in BIC.
  - If the modified network structure improves BIC, accept, and update the network structure (reject otherwise).
  - Terminate the loop if BIC is repeating more than twice.
- Ensure the global connectivity of the network using the depth-first search (DFS) algorithm [53].
- If the network is not globally connected, re-initialize the random network structure.
- If the functional connectivity network is globally connected, save the connectivity matrix, and shift to the next time window.

A time sliding window of length $L$ seconds on the EEG time series signals is used in this study. The sliding window approach helps in capturing the dynamic behavior of the EEG signals and is used in different studies in literature [54], [55]. For each sliding window, the functional connectivity matrix using BSL is estimated. The sliding window is moved forward to estimate the functional connectivity [56], using the RICI algorithm [57]. Finally, the dynamic connectivity matrix of shape $N*N*L$ is extracted.

#### II. BAYESIAN INFORMATION CRITERION

Let G = (V,E) be a directed acyclic graph (DAG) that represents a set of random variables $X_1, X_2, ..., X_n$, where each node in the graph corresponds to a variable and each edge represents a direct causal relationship between two variables. Bayesian score of this network structure is determined, given a network structure, reflects the likelihood of the observed data, considering the complexity of the structure [42]. Let

$$D = \{D_1, D_2, ... D_m\} \quad (1)$$

be a set of m observed data points, where,

$$D_i = \{x_1^i, x_2^i, ..., x_n^i\} \quad (2)$$

is a vector of n observed values for each variable in the network, corresponding to the ith data point. Given the network structure (assuming that data points are independent and identically distributed).

The Bayesian score of the network structure $G$ is defined as:

$$score(G|D) = p(D|G)p(G) \quad (3)$$

Where the likelihood of the data given the network structure is denoted by $p(D|G)$, and prior probability of the network structure is denoted by $p(G)$.

The likelihood function can be expressed as a product of conditional probabilities using the chain rule of probability [58]:

$$p(D|G) = \prod_{i=1}^{m} p(D_i|G) = \prod_{i=1}^{m} \prod_{j=1}^{n} p(x_j^i | Pa_j^i, \theta_j) \quad (4)$$

where $Pa_j^i$ are the parents of variable Xj in the network, and $\theta_j$ is the parameter of the conditional probability distribution of $X_j$ given its parents $Pa_j^i$.

The conditional probability distributions of the variables are parameterized by hyperparameters α and β, which reflect the prior beliefs or uncertainties about the distribution. Specifically, Dirichlet and conjugate Gaussian priors [59] are used:

$$p(\theta_j|\alpha) = Dirichlet(\theta_j|\alpha) \quad (5)$$

$$p(x_j|Pa_j, \theta_j, \beta) = N(x_j|f(Pa_j, \theta_j, \beta^{-1})) \quad (6)$$

where $f(Pa_j, \theta_j)$ is a function that maps the parent values to the expected value of $x_j$, given the parameters $\theta_j$, N represents normal (Gaussian) distribution.

The network structure G is a priori equally likely to be any DAG over the same set of nodes, so the prior probability is uniform:

$$p(A) = \frac{1}{A} \quad (7)$$

where $A$ is the total number of possible DAGs over n nodes.

The data observed are used to calculate the estimated conditional probabilities of node *i* given its parent nodes, indicative of a multinomial distribution.

Then use a Dirichlet prior distribution with hyperparameters $α_1, α_2, ..., α_k$ to model the prior distribution over the conditional probabilities.

The score for node i can be computed using the equation 8:

$$score(i) = \log P(m_i|\alpha) + \log P(\alpha) - \log P(m_i, \alpha) \quad (8)$$

where $m_i$ is the observed count of node i. $\alpha$ is the hyperparameter for Dirichlet prior distribution. $P(m_i|\alpha)$ is the likelihood of observing the data $m_i$ given the hyperparameters $\alpha$. It quantifies how probable the observed counts are under the prior beliefs. The joint distribution of the observed $m_i$ and the hyperparameters $\alpha$ is represented by $P(m_i, \alpha)$.

Simplifying equation 8:

$$P(m_i|\alpha) = \int P(m_i|\theta_i) P(\theta_i|\alpha) d\theta \quad (9)$$

Represent the prior distribution over $\theta_i$ using the Dirichlet distribution:

$$P(\theta_i|\alpha) = Dir(\theta_i|\alpha) \quad (10)$$

where α is the hyperparameter vector for the Dirichlet distribution. The Dirichlet distribution has the following properties:

$$\int Dir(\theta_i|\alpha) d\theta_i = 1 \quad (11)$$

$$E[\theta_i] = \frac{\alpha_i}{\sum_p \alpha_p} \quad (12)$$

Using equation (11) and equation (12) and the properties of the Dirichlet distribution, simplify the expression for score(i) as follows:

$$score\ (i) = \log \int P(m_i|\theta_i)P(\theta_i|\alpha)d\theta + \log \prod_p P(\alpha_p) - \log \int P(m_i|\theta_i)P(\theta_i|\alpha)P(\alpha)d\theta_i\ d\alpha \quad (13)$$

Finally,

$$score\ (i) = \log \Gamma\left(\sum_p \alpha_p\right) - \sum_p \log \Gamma(\alpha_p) + \sum_p \log \Gamma(\alpha_p + m_i, p) - \log \Gamma\left(\sum_p (\alpha_p + m_i, p)\right) \quad (14)$$

Where gamma function is denoted by Γ(x), the properties of the gamma function and the Dirichlet distribution are used to simplify the expression. The observed count is $m_i$,p, of the joint occurrence of node i and its parent $j_p$ in the data.

The resulting equation for *score(i)* represents the log-likelihood of the data given the conditional probabilities of node i and its parents. This equation is used to compute the score for each node (EEG channel) in the graph and the total score for the entire graph.

### C. Classification

III. GRAPH CONVOLUTIONAL NETWORK (GCN):

The proposed GCN model can be divided into three sequential modules: input, ResNet, and classification. The architecture of the proposed model is displayed in Figure 2. The functional connectivity networks are given as an input to the GCN model. k-fold cross-validation is performed, where we use (k-1) folds for training and validating the model, and one-fold is used for testing the model. Each module is discussed in this section.

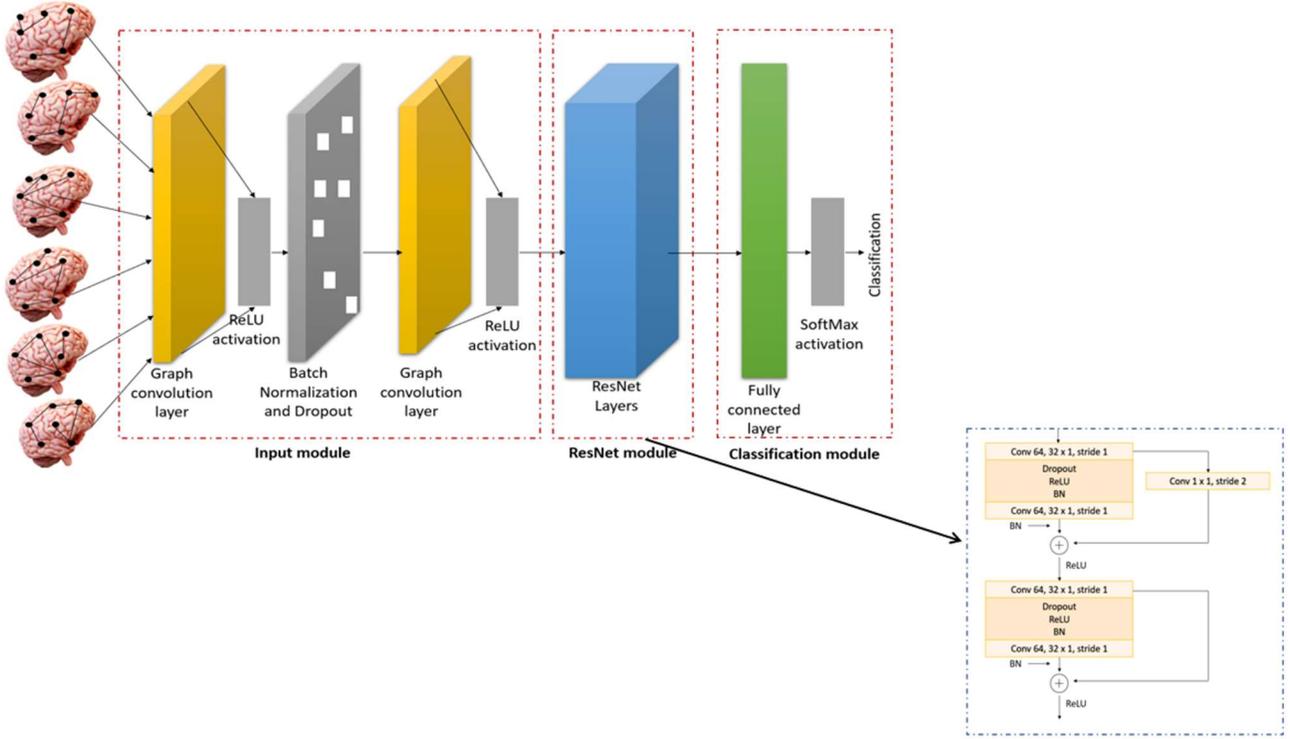

Figure 2. Architecture of GCN. The model consists of input module, ResNet module, and classification module. Functional connectivity networks are fed into the input module that has graph convolution layers. The ResNet module consists of convolutional block and identity block with a "shortcut" structure.

*1) Input module:* The input module is a graph convolutional module that can efficiently handle irregular and non-uniform data. Moreover, it can capture both local and global structures of graphs by incorporating information from neighboring nodes and their edges. This allows GCN to learn representations that capture both local and global properties of the underlying graph. This module has two graph convolutional layers with a symmetric adjacency matrix that is weighted with edge attributes [60].

$$z = f(X, A) = (\hat{A} \, Re \, LU \, (\hat{A} \times w^{(0)})w^{(1)}) \quad (15)$$

where,

$$\hat{A} = \widetilde{D}^{-\frac{1}{2}} \widetilde{A} \widetilde{D}^{-1/2} \quad (16)$$

Here, X is a feature matrix of N*D dimensions where N represents number of nodes, and D represents number of input features, A represents the weighted (or sparse) adjacency matrix of the graph. The outcome is a node-level output Z, which is an N*F feature matrix where F represent the number of features for each node. The hidden neural network can be written as a non-linear function:

$$H^{(l+1)} = f(H^{(l)}, A) \quad (17)$$

With $H^{(0)} = X$ and $H^{(L)} = Z$, where L is the number of layers. $(\hat{A} \, Re \, LU \, (\hat{A} \times w^{(0)})w^{(1)})$ is non-linear activation function of ReLU. In equation (15), $w^{(0)}$ serves as the transformation matrix from input to the hidden layer, accommodating H feature maps within that hidden layer. The weight matrix $w^{(1)}$ functions as the transformation from the hidden layer to the output.

*2) ResNet module:* Deep CNN layers often encounter gradient problems, resulting in network degradation. The issue of network degradation can be mitigated by implementing a shortcut structure that learns features through the residual, which is the difference between the optimal mapping (H(x)) and the identity mapping (x).

$$F(y) = H(x) - x \quad (18)$$

The ResNet module contains a convolutional block and an identity block (Figure 2). In addition, a dropout and batch normalization (BN) layer are added to avoid overfitting and the 'internal covariate shifting' respectively [61].

*3) Classification block:* The classification block is composed of a fully connected layer followed by a classification layer that employs a Softmax activation function.

## III. RESULTS

The model operates on a PC equipped with an Intel® Core™ i7 processor, 16 GB of RAM, and an NVIDIA GeForce RTX 2070 GPU with 8 GB of memory, utilizing Python 3.7 and TensorFlow 2.2 as the software framework.

### A. Performance of the proposed method with frequency bands:

Alpha, beta, and theta oscillatory synchrony play a significant role in WM tasks by activating the task-relevant cortical areas. Furthermore, beta and theta synchronizations plays a crucial role in controlling the flow of information in verbal WM tasks [62], [63], [64], [65], [66], [67], [68], [69]. The functional connectivity between EEG channels in alpha band is displayed in Figure 3(a). For easy understanding of functional connectivity, the graph is plotted for edge strength ($\varepsilon$)>1.5. The connectivity graph in Figure 3 (a) also displays the higher connectivity in Frontal and Parietal regions for 5M that is aligning with the alpha band activations of topoplots. Higher connectivity in central region is observed ($\varepsilon$>1.2) in beta band that are aligning with the beta activations that can be found in Figure 3 (b). Lower power and lower connectivity is observed for 7M in beta band. Higher connectivity ($\varepsilon$>1.5) is observed for manipulation tasks compared to retention tasks in theta band in the frontal midline region (see Figure 3(c)). The functional connectivity metrics showed consistent reproducibility in alpha and theta bands in comparison with more granular beta. Also, weaker connections in the beta band are observed with increasing memory loads. The BSL algorithm produced consistent results in all three frequency bands compared to other functional connectivity metrics. Verbal and non-verbal WM stimuli were compared in [63], showing enhanced power in the theta band when compared to alpha and beta bands. Alpha-beta decrements have been observed in ADHD adolescents [64] while performing WM tasks. Memory load increased the mean AEC-c in alpha and beta bands in MCI patients compared to healthy subjects in visuospatial WM maintenance [65]. A comparison between manipulation and retention tasks in [66] showed increased power in frontal lobe in theta band. All the above studies confirm the behavior of the BSL algorithm in estimating functional connectivity in alpha, beta, and theta frequency bands.

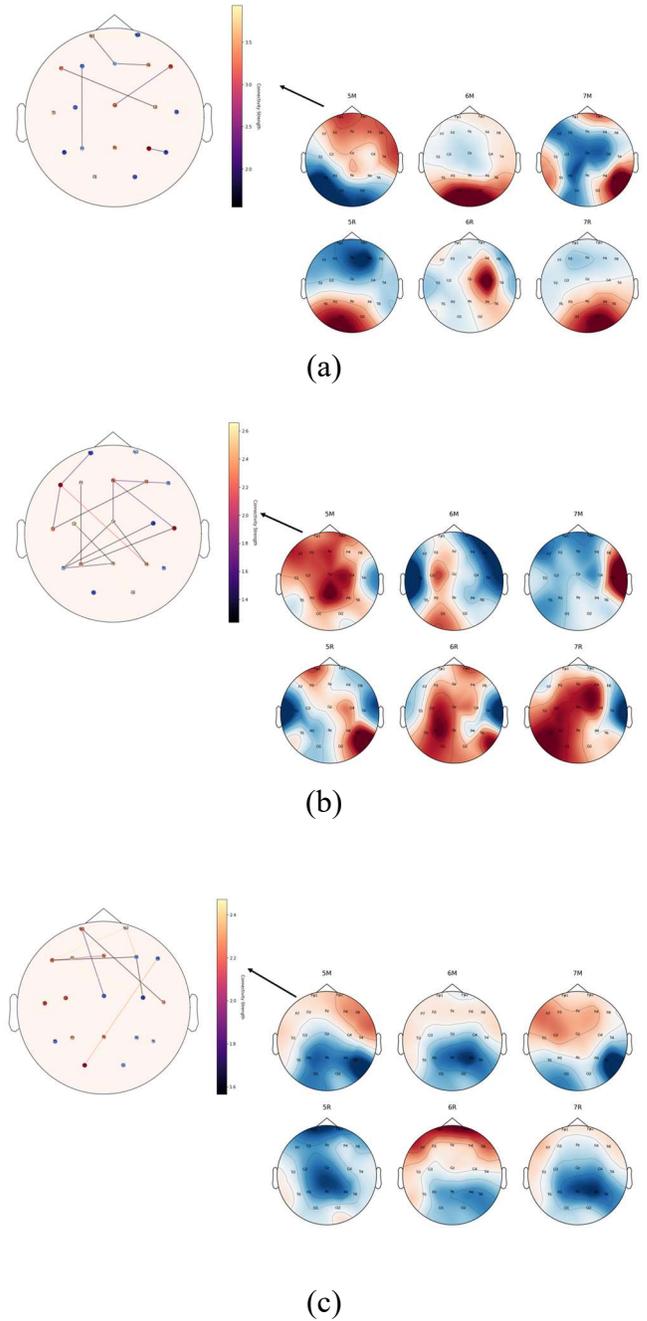

*Figure 3. (a) Topoplots for alpha band for different memory and cognitive loads. The figure also displays functional connectivity graph for edge strength ($\varepsilon$)>1.5 displaying higher connectivity in frontal and parietal region in alpha band for 5M. (b) Topoplots for beta band for different memory and cognitive loads. ) The figure also displays functional connectivity graph for edge strength ($\varepsilon$)>1.2 displaying higher connectivity in central region for beta band. (c) Topoplots for theta band for different memory and cognitive loads. The figure also displays functional connectivity graph for edge strength ($\varepsilon$)>1.5 displaying higher connectivity in frontal midline region in theta band. (5M) memory*

*load 5, cognitive task manipulation. (6M) memory load 6, cognitive task manipulation. (7M) memory load 7, cognitive task manipulation. (5R) memory load 5, cognitive task retention. (6R) memory load 6, cognitive task retention. . (7R) memory load 7, cognitive task retention.*

### B. *Statistical analysis of variance (ANOVA)*

Intersubject and intrasubject one-way ANOVA for the cognitive and memory loads (5M, 6M, 7M, 5R, 6R, 7R) is performed. The statistical significance for all analyses was at p-value < 0.05. The ANOVA results indicate significant differences in EEG frequency band powers both within and between subjects. Specifically, the beta and theta bands show statistically significant differences across different conditions within subjects (intrasubject), with p-values of 0.012 and 0.0394, respectively. There are also significant differences in the power of these bands between subjects (intersubject), as evidenced by p-values of 0.0072 for beta and <0.005 for theta. For the alpha band, a significant intersubject difference is observed (p<0.001), but the intrasubject difference borders on significance (p=0.059).

### C. *Reproducibility of single-subject inference*

The correlations of edge strengths of connectivity metrics for different trials of the same WM loads of the same subject data are compared to find out whether the metrics produce similar functional connectivity matrices. Spearman correlation is used to compare the connectivity matrices for different trials of the same WM loads. Spearman correlation gives us an insight into the reproducibility of functional connectivity matrices of the same subject for different trials of the same cognitive load [70]. The comparison of BSL Spearman correlation with state-of-the-art functional connectivity metrices is presented in Figure 4. The Spearman correlation produced comparable results. BSL had a consistent correlation between different trials of the same WM loads, of which the alpha band has the highest correlation (0.972) followed by theta band (0.90). The beta band has the lowest correlation for BSL (0.81). However, BSL had more consistent results in all three frequency bands compared to other functional connectivity metrices.

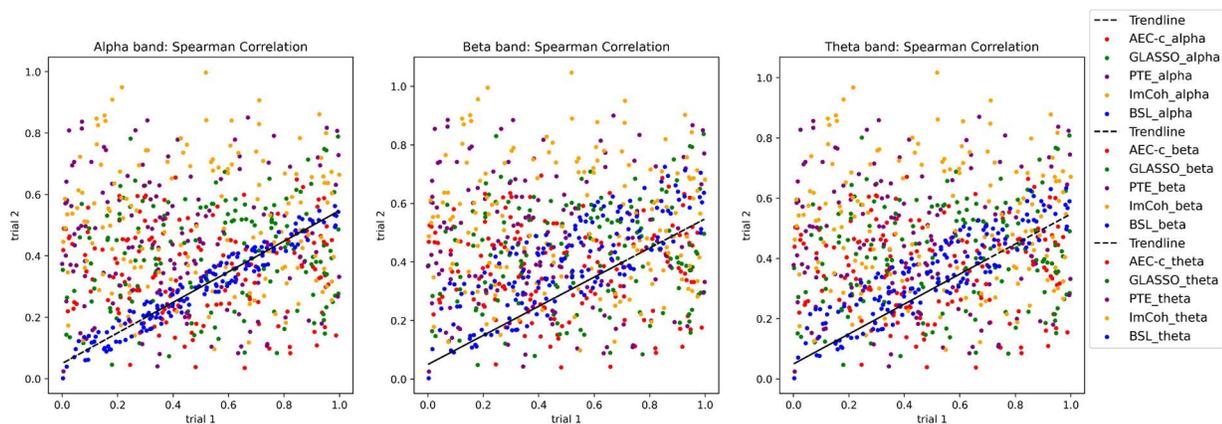

*Figure 4. Comparison of BSL Spearman correlation with state-of-the-art functional connectivity metrices*

### D. Intrasubject classification

A six-class intrasubject (subject-specific) classification is performed by training, testing, and validating the model for each subject. The accuracy, sensitivity, specificity, and kappa coefficients of the top 10 classifications in alpha, beta, and theta bands are presented in Table 2. The confusion matrices and precision-recall curves can be provided on-demand for reference. The best classification accuracy in the alpha band is 96.67% (sub-16), in the beta band is 86.7% (sub-66), and in theta band is 92.7% (sub-5). Sub-3 is one of the top 10 classifications in all frequency bands. The Kappa coefficient measures the inter-rater reliability [71]. Across all frequency bands, the kappa coefficients suggest that the results are not only accurate but also consistently reliable. The range of classification accuracy range within alpha band is 87-96%, beta band is 74-86%, theta band is 89-92%. Classification in alpha and theta bands is consistently high compared to the beta band.

The sensitivity and specificity results provide a more granular understanding of the model performance across the individual classes. For all frequency bands, the sensitivity and specificity values suggest that the classifier performs well in identifying positive cases (high sensitivity) and negative cases (high specificity). In the alpha band, performance seems to be robust, with both sensitivity and specificity values being above 90% for the top 10 classifications.

*Table 1. Accuracy, sensitivity, specificity, and kappa coefficient of top 10 classifications in alpha, beta, and theta bands*

| Frequency bands | Subject | Accuracy | Sensitivity | Specificity | Kappa coefficient |
|---|---|---|---|---|---|
| Alpha band | Sub-16 | 0.967 | 0.967 | 0.993 | 0.960 |
| | Sub-8 | 0.960 | 0.960 | 0.992 | 0.952 |
| | Sub-3 | 0.953 | 0.953 | 0.991 | 0.944 |
| | Sub-5 | 0.947 | 0.947 | 0.989 | 0.936 |
| | Sub-15 | 0.940 | 0.940 | 0.988 | 0.928 |
| | Sub-27 | 0.933 | 0.933 | 0.987 | 0.920 |
| | Sub-2 | 0.927 | 0.927 | 0.985 | 0.912 |
| | Sub-35 | 0.920 | 0.920 | 0.984 | 0.904 |
| | Sub-6 | 0.913 | 0.913 | 0.983 | 0.896 |
| | Sub-9 | 0.907 | 0.907 | 0.981 | 0.888 |
| Beta band | Sub-66 | 0.867 | 0.867 | 0.973 | 0.840 |
| | Sub-4 | 0.860 | 0.860 | 0.972 | 0.832 |
| | Sub-37 | 0.853 | 0.853 | 0.971 | 0.824 |
| | Sub-38 | 0.847 | 0.847 | 0.969 | 0.816 |
| | Sub-11 | 0.840 | 0.840 | 0.968 | 0.808 |
| | Sub-17 | 0.833 | 0.833 | 0.967 | 0.800 |
| | Sub-23 | 0.827 | 0.827 | 0.965 | 0.792 |
| | Sub-14 | 0.820 | 0.820 | 0.964 | 0.784 |
| | Sub-3 | 0.813 | 0.813 | 0.963 | 0.776 |
| | Sub-1 | 0.807 | 0.807 | 0.961 | 0.768 |
| Theta band | Sub-5 | 0.927 | 0.927 | 0.985 | 0.912 |
| | Sub-13 | 0.920 | 0.920 | 0.984 | 0.904 |
| | Sub-18 | 0.913 | 0.913 | 0.983 | 0.896 |
| | Sub-35 | 0.907 | 0.907 | 0.981 | 0.888 |
| | Sub-25 | 0.900 | 0.900 | 0.980 | 0.880 |
| | Sub-26 | 0.893 | 0.893 | 0.979 | 0.872 |
| | Sub-3 | 0.887 | 0.887 | 0.977 | 0.864 |
| | Sub-15 | 0.880 | 0.880 | 0.976 | 0.856 |
| | Sub-37 | 0.873 | 0.873 | 0.975 | 0.848 |
| | Sub-14 | 0.867 | 0.867 | 0.973 | 0.840 |

### E. Performance of the proposed method with different features (functional connectivity metrics)

State-of-the-art machine learning algorithms have used various functional connectivity features imposing amplitude, spectral, and phase synchronizations. To investigate the efficiency of the proposed model, different functional connectivity features are extracted that are corrected for spatial leakage in literature, including Amplitude envelope correlation (AEC-c), Imaginary coherence (ImCoh) [72], and Phase Transfer Entropy (PTE) [73]. The proposed methodology is also compared with graphical LASSO. These functional connectivity features are classified independently using the proposed GCN model for accuracy. Table 2 presents the performance metrics of the proposed model. BSL, AEC-c, and graphical LASSO (GLASSO) [74] obtain accuracies over 90% as does the proposed classification model for the subject with the highest classification accuracy among all subjects.

*Table 2. Performance comparison of functional connectivity metrics using the proposed classification model (GCN)*

| Features | Accuracy (%) (Best classification) | Accuracy (%) Average of all subjects |
|---|---|---|
| AEC-c | 93±0.28 | 82±0.32 |
| ImCoh | 77.9±0.25 | 63±0.16 |
| PTE | 86.11±1.26 | 75±0.45 |
| Graphical LASSO | 91.2±0.84 | 79±0.67 |
| **BSL (proposed)** | **96.67±0.19** | **89±0.52** |

### F. Comparison of the proposed BSL feature extraction method with different classifiers

The competence of the proposed BSL algorithm is investigated using different classifiers proposed in the literature. These include SVM, CNN, k-NN, and LDA. The classifiers SVM and LDA are designed for binary class classification. However, multiple SVM or LDA classifiers are used in a one-vs-one manner to perform a multiclass classification, resulting in increasing computational cost [75], [76]. Therefore, each of these classifiers in a one-vs-one manner is combined using the Python module Scikit-Learn module . k-NN is also used from the Scikit-Learn module in Python (version 3.6). CNN is also used from using TensorFlow (version 2.2) according to the proposed architecture in [48]. Table 3 presents the performance of the proposed method with other classifiers. The best classification accuracy of the proposed GCN model is 96% alpha band. The average classification accuracy of all subjects with the GCN model is 89%, which also outperformed other classifiers. SVM has the second-highest classification accuracy for both the best subject and the average accuracy of all subjects.

*Table 3. Performance comparison of the proposed BSL functional connectivity classification with other classifiers*

| Classifier | Accuracy (%) Best classification | Accuracy (%) Average of all subjects |
|---|---|---|
| LDA | 68±2.98 | 61±2.98 |
| SVM | 92.46±0.28 | 80±0.28 |
| CNN | 85.9±0.25 | 78±0.25 |
| k-NN | 69.2±2.74 | 66±2.74 |
| **proposed model** | **96±0.19** | **89±0.52** |

### G. Comparison of proposed methods with state-of-the-art methods

The proposed method achieved a classification accuracy of 96% which outperformed state-of-the-art methods (Table 4). The research on classifying cognitive loads using functional connectivity metrics is very limited. In addition, the classification of EEG verbal WM dataset is performed for the first time and hence cannot compare our results with state-of-the-art results. Therefore, these results are the benchmark for further comparisons on this dataset. However, we have compared state-of-the-art methods that have used functional connectivity features for classification. Table 4 presents the comparison of those methods. The functional connectivity metrics used in the literature are derived from functional connectivity approaches that infer amplitude, spectral, and phase synchronizations. These methods draw spurious connections among the EEG channels, and a threshold is used to derive the functional connectivity metrics. Therefore, such connections draw confounding metrics that lead to low classification accuracies of the models.

*Table 4. Performance comparison of the proposed method (BSL + GCN) with state-of-the-art methods*

| Paper | Working Memory task | Features | Classifier | Accuracy (%) |
|---|---|---|---|---|
| Xi J et. al. [43] | memory | PLV | SVM | 78 |
| Dimitrakopoulos et. al. [44] |  | Pearson correlation | SVM | 88 |
| Zhang et. al. [77] | n-back task | Graph topology | Two-stream neural network | 88.9 |
| Gupta et. al. [48] | n-back task | PLV | CNN-LSTM | 93.75 |
| Zhang et. al. [78] | Paired associates learning | Graph topology | CNN (modified) | 91 |
| Gupta et. al. [46] | n-back | PTE | BrainNetCNN | 92 |
| Liu et. al. [79] | Sternberg WM task | Coh, power spectral density, and microstates | SVM | 84.49 |
| Proposed method | Verbal WM task | BSL | GCN | 96 (highest classification accuracy among all subjects ) 89 (average of all classification accuracies) |

### H. Male Vs Female performance analysis

The verbal WM EEG data was collected from 81 female subjects and 73 male subjects. The accuracy of the performance of the WM tasks is measured and compared in female subjects and male subjects (Figure 5). The performance analysis is correlated with the functional connectivity networks where high performance of both male and female subjects is observed in the retention task in alpha and theta bands. The performance of female subjects is higher in manipulation tasks compared to the male subjects. Pearson's correlation reveals that both male ($r = 0.14$) and female ($r = 0.40$) subjects have a high correlation in memory load 5 between manipulation and retention tasks. The least correlation of cognitive tasks in females is with memory load 7 ($r = 0.24$) whereas in males, it is with memory load 6 ($r = 0.07$).

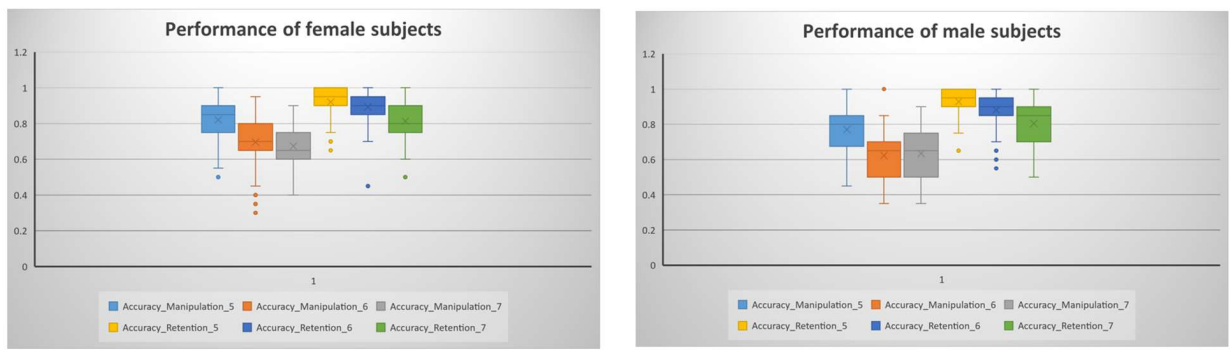

*Figure 5. Performance analysis of male vs female subjects*

## IV. DISCUSSION

The current study presents a novel approach using BSL and GCN to classify WM load from EEG data. The application of BSL has demonstrated a significant enhancement over traditional EEG functional connectivity methods in sensor space, providing a refined approach for capturing subtle changes in WM load and significantly boosting classification accuracy (see Table 2). Our results indicate that the subject-specific functional connectivity features, harnessed through the GCN, lead to a substantial increase in classification performance when compared with state-of-the-art methods (see Table 4).

The BSL algorithm produced consistent results across alpha, beta, and theta frequency bands and highlights its robustness and adaptability to the complex nature of EEG data. The topoplots for alpha band activations (see Figure 3) demonstrate increased power in the frontal and parietal regions, consistent with the established role of the alpha band in WM tasks [80], [81]. This band, traditionally linked to inhibitory control and attentional processes [82], shows a pronounced activation that correlates with task-relevant cortical areas. The increased activity in these regions during WM tasks is indicative of the role of alpha band in modulating attention and suppressing irrelevant information, which is critical for maintaining focus on the task [83].

The beta band is known for its association with active processing and maintenance of cognitive states [84]. Our analysis shows that while performance of the beta band in classification accuracy is not as high as alpha or theta bands, it still plays a crucial role in the flow of information during WM tasks [84]. Higher activations in central regions are observed in beta band, which is closely intertwined with cognitive processing [85]. During WM tasks, especially those requiring a motor response, the beta band may facilitate the translation of cognitive decisions into motor actions [86]. This aligns with existing literature that associates beta activity with motor control and planning.

The robust performance in classification within the theta band is consistent with the established role of this frequency in modulating neural activities crucial for memory processes [87]. The theta band's involvement is particularly crucial in tasks that require coordination of neural activity across different brain regions, which is essential for complex cognitive functions.

The distinct patterns observed in these frequency bands offer an understanding of the functional dynamics of the brain during WM tasks. Further research should focus on integrating these findings with multimodal neuroimaging data to gain a more comprehensive understanding of the brain's functional connectivity. Additionally, exploring the interplay between these frequency bands in different cognitive states or disorders could provide deeper insights into the neural basis of cognitive functions and dysfunctions.

The reproducibility of single-subject inference supported by high Spearman correlation coefficients further confirms the reliability of BSL method in capturing functional connectivity (see Figure 4). Comparisons with different features and classifiers have demonstrated the consistency of BSL method. When compared with traditional classifiers such as SVM, CNN, k-NN, and LDA, the GCN model showcased the best performance, corroborating the effectiveness of integrating machine learning techniques with neuroscientific research (see Table 4). The competence of the model was further substantiated through rigorous 10-fold cross validation, ensuring the validity of our results.

Analyzing the gender differences in cognitive performance provides an insight into how biological and sociocultural factors influence brain functioning and cognitive abilities [88]. The current study revealed notable differences in performance between male and female subjects while performing WM tasks. Specifically, the female subjects exhibited higher performance in manipulation tasks compared to male subjects. Recognizing gender differences in cognitive abilities is crucial to develop effective neurophysiological assessments and interventions [89]. The observed differences prompt further investigation into underlying mechanisms driving gender variations. Future research should aim to explore the neurobiological underpinnings of these differences, by integrating multimodal neuroimaging. Additionally, more research is necessary to better understand cognitive development and aging in both genders [90].

## V. Conclusion

In this manuscript, a BSL method to estimate the dynamic functional connectivity for EEG verbal WM is presented. This approach outperforms the traditional EEG functional connectivity methods in sensor space. The classification model is built on subject-specific functional connectivity features using GCN. The subject-specific model performance is compared with state-of-the-art functional connectivity features with 10-fold crossvalidation. The BSL functional connectivity features efficiently captured the subtle changes in WM load and increases the classification accuracy significantly. The proposed GCN model give promising results for not only BSL features, but also for other traditional functional connectivity metrics. The selection of frequency bands displayed significant impact on performance measures such as accuracy, sensitivity, specificity, ANOVA, and Spearman correlation. Although the behavioral results suggest that the performance of female subjects is better than male subjects, further investigation is needed on the functional connectivity patterns of male vs female subjects.